\documentclass[letterpaper]{article} 
\usepackage{aaai25}  
\usepackage{times}  
\usepackage{helvet}  
\usepackage{courier}  
\usepackage[hyphens]{url}  
\usepackage{graphicx} 
\usepackage{natbib}  
\usepackage{caption} 
\usepackage{amsmath}
\usepackage{amsfonts}
\usepackage{algorithm}
\usepackage{algorithmic}
\usepackage[table]{xcolor}
\usepackage{multirow} 
\usepackage{booktabs}
\usepackage{newfloat}
\usepackage{listings}
\DeclareCaptionStyle{ruled}{labelfont=normalfont,labelsep=colon,strut=off} 
\floatstyle{ruled}
\newfloat{listing}{tb}{lst}{}
\floatname{listing}{Listing}
\title{ Exploiting Multimodal Spatial-temporal Patterns for Video Object Tracking}
\author {
    Xiantao Hu\textsuperscript{\rm 1},
    Ying Tai\textsuperscript{\rm 2,1}\thanks{ Ying Tai and Jian Yang are the corresponding authors.},
    Xu Zhao\textsuperscript{\rm 1},
    Chen Zhao\textsuperscript{\rm 2},
    Zhenyu Zhang\textsuperscript{\rm 2},
    Jun Li\textsuperscript{\rm 1},
    Bineng Zhong\textsuperscript{\rm 3},
    \\
    Jian Yang\textsuperscript{\rm 1}\footnotemark[1]
}
\affiliations {
    \textsuperscript{\rm 1}PCA Lab, Key Lab of Intelligent Perception and Systems for High-Dimensional Information of Ministry of Education, School of Computer Science and Engineering, Nanjing University of Science and Technology\\
    \textsuperscript{\rm 2}Nanjing University 
    \textsuperscript{\rm 3}Guangxi Normal University\\
    
    \{xiantaohu, csjyang, junli\}@njust.edu.cn, \{yingtai,zhenyuzhang\}@nju.edu.cn, bnzhong@gxnu.edu.cn
}

\usepackage{bibentry}

\begin{document}

\maketitle

\begin{abstract}
Multimodal tracking has garnered widespread attention as a result of its ability to effectively address the inherent limitations of traditional RGB tracking. However, existing multimodal trackers mainly focus on the fusion and enhancement of spatial features or merely leverage the sparse temporal relationships between video frames. These approaches do not fully exploit the temporal correlations in multimodal videos, making it difficult to capture the dynamic changes and motion information of targets in complex scenarios. To alleviate this problem, we propose a unified multimodal spatial-temporal tracking approach named STTrack. In contrast to previous paradigms that solely relied on updating reference information, we introduced a temporal state generator (TSG) that continuously generates a sequence of tokens containing multimodal temporal information. These temporal information tokens are used to guide the localization of the target in the next time state, establish long-range contextual relationships between video frames, and capture the temporal trajectory of the target. Furthermore, at the spatial level, we introduced the mamba fusion and background suppression interactive (BSI) modules. These modules establish a dual-stage mechanism for coordinating information interaction and fusion between modalities. Extensive comparisons on five benchmark datasets illustrate that STTrack achieves state-of-the-art performance across various multimodal tracking scenarios. 
Code is available at: \url{https://github.com/NJU-PCALab/STTrack}.
\end{abstract}


%
\begin{figure}[t]
  \centering
    \includegraphics[width=1\linewidth,height=6.8cm]
    {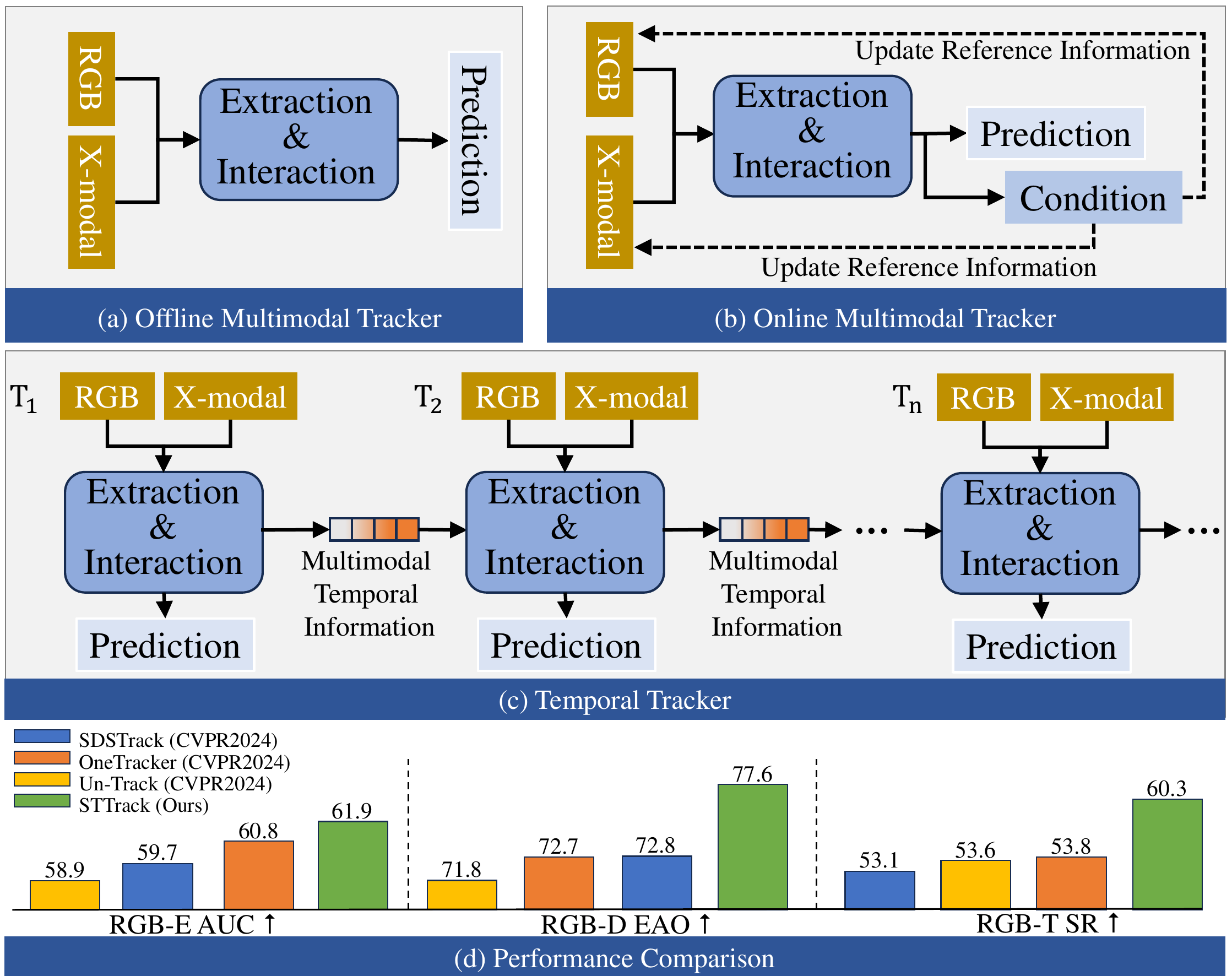}
   \caption{
\textbf{Illustrations of different frameworks of multimodal trackers (a)-(c), and performance comparison (d)}. (a) Offline multimodal tracker performs offline tracking of video sequences using fixed template frames.
 (b) Online multimodal tracker is based on an updating strategy, which utilizes the results condition to update the reference information.
 (c) Our proposed STTrack transmits multimodal temporal information throughout the tracking process. 
 (d) STTrack achieves superior performance against recent state-of-the-art competitors on three popular multimodal tasks.}
   \label{fig:contrast}

\end{figure}
  
\section{Introduction}
Visual object tracking is the process of locating and following a specific object across consecutive frames in a video sequence. As a fundamental vision task, it is essential for various applications~\cite{wang2022learning,wang2025gpsformer,jiang2023lttpoint, zhang2024deformation} and their related tasks~\cite{anSHaRPoseSparseHighResolution2024,zheng2023curricular,zhang2024few,fang2023multi,nan2024openvid,ning2023pedestrian}.
Despite numerous efficient RGB-based trackers~\cite{AQAtrack,evptrack,FFtrack,artrack,siamban,transt,siamban_p,xue2024unifying} have been proposed through high quality dataset~\cite{lasot,got10k,trackingnet}, they are still limited by the degradation of RGB imaging quality caused by the complexity of real-world scenarios, which leads to tracking errors.
Compared to RGB modalities, thermal infrared (TIR) provides clear target information in low light environments; depth modalities offer distance cues from depth cameras; and event modalities use event-based cameras to capture motion information and generate stable target trajectories.
Therefore, developing an effective multimodal tracker that combines various modality X (such as TIR, depth, and event) with RGB is crucial for robust tracking.

Multimodal tracking methods can be broadly categorized into: \textit{Offline trackers with fixed template frames} and \textit{online trackers that update reference information}. 
$1$) Traditional offline multimodal trackers focus on the fusion and interaction of spatial multimodal features, evolving from early CNN architectures~\cite{protrack,apfnet,mfDiMP} to the recent Transformer architectures~\cite{chen2024top,mctrack}. 
As shown in Fig.~\ref{fig:contrast}~(a), offline trackers rely on a fixed initial target appearance as reference information for the entire tracking process. 
However, as time passes, the target may deform or become occluded, rendering the initial template frame unable to accurately capture its current state.
$2$) In contrast, as depicted in Fig.~\ref{fig:contrast}~(b), online trackers capture more recent target appearance features by updating reference information, such as template images~\cite{tatrack,mplt}, search images~\cite{TAAT}, or historical frame features~\cite{DMSTM}. 
Although these approaches enable updates at specific points in time, their reliance on sparse temporal relationships (\textit{i.e.}, updates limited to specific conditions) neglects the continuity of temporal information. 
In video tracking tasks, \textit{target changes and movements} typically follow a certain trend, which is challenging to capture and express without explicit temporal modeling, thus limiting the model's performance in complex scenarios.

To address this issue, we propose a novel tracking framework STTrack based on multimodal spatial-temporal patterns. 
STTrack improves to capture and represent the dynamic target by explicitly leveraging the temporal context within multimodal video data.
As shown in Fig.~\ref{fig:contrast}~(c), we make full use of the multimodal temporal information from videos to guide the modeling of the current state of targets, thereby constructing a unified multimodal temporal strategy. 
There are several critical modules in STTrack. 
$1$) At the temporal level, we design a novel \textit{temporal state generator (TSG) based on cross mamba architecture}~\cite{sigma}. 
TSG combines the current cross-modal target representation features with previous multimodal temporal information, employing an autoregressive mechanism to generate multimodal temporal information tokens for the current time step.
These tokens act as bridges for information transfer, facilitating the tracking process for the next time node. 
$2$) At the spatial level, since cross-modal interaction and fusion are crucial for effective multimodal tracking,
we therefore propose the \textit{background suppression interactive} module in the feature extraction stage of the visual encoder, and the \textit{mamba fusion} module in the final modality fusion stage, respectively. 
The BSI module improves each modality branch’s representation by integrating features from other modalities, while the mamba fusion module dynamically merges multimodal features from both branches to facilitate precise object localization.

We summarize the contributions of STTrack as follows:

\begin{itemize}
\item To fully exploit the temporal information from multiple modalities, we propose STTrack, which introduces temporal state generator to reveal temporal context of target.

\item We propose the BSI and mamba fusion modules, which optimize information interaction and dynamic fusion between modalities during the feature extraction and modality fusion stages, respectively.
\item The proposed STTrack achieves state-of-the-art performance on five popular multimodal tracking benchmarks, including RGBT234, LasHeR, VisEvEnt, Depthtrack, and VOT-RGBD2022.
\end{itemize}

\begin{figure*}
  \centering
    \includegraphics[width=1\linewidth,height= 6 cm]
    {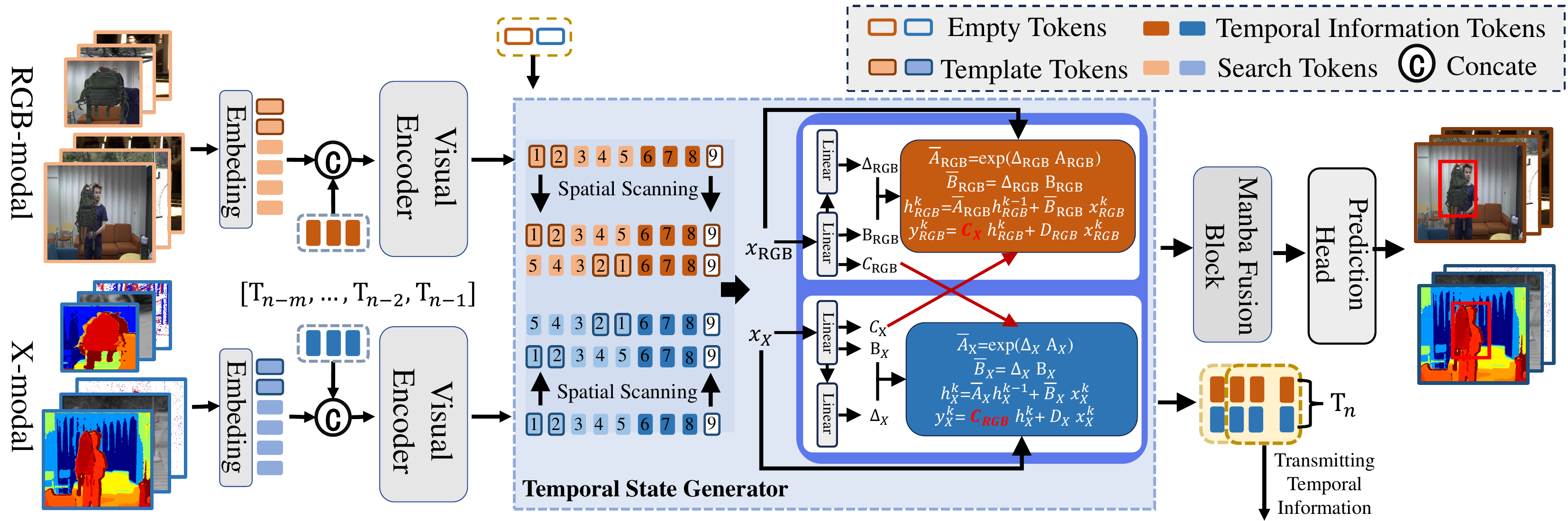}
   \caption{
\textbf{Overall architecture of STTrack}. The temporal information tokens of each modality, along with the image tokens, are fed into the vision encoder to guide the extraction of current features using temporal information. In our designed Temporal State Generator, the current temporal tokens are generated based on cross-modal features and previous temporal features. We have added cross modal interaction in Visual Encode. Finally, the features are finely adjusted and fused through the mamba fusion module and then fed into the tracking head to predict the current state.}
   \label{fig:structure}
\end{figure*}

\section{Related Works}
\textbf{Multimodal Tracking.} 
In recent years, multimodal tracking has gained widespread attention for its ability to achieve robust tracking in complex scenarios. By allowing different modalities to complement each other, it overcomes challenges that a single modality cannot address on its own.
Early multimodal tracking methods~\cite{ADRNet,CAT} typically focused on specific multimodal task. 
For instance, APFNet introduces the concept of attribute fusion based on ResNet~\cite{resnet}, enhancing its performance under specific challenges. 
TBSI~\cite{tbsi} extends ViT~\cite{vit_transformer} to RGB-T tasks and leverages the TBSI module to optimize cross-modal interactions. 
More recently, some works~\cite{oneTracker} have begun exploring unified architectures capable of handling multiple multimodal tasks. 
ViPT~\cite{vipt} integrates other modalities into the RGB modality through a prompt mechanism. 
SDSTrack~\cite{sdstrack} and BAT~\cite{bat} explore symmetrical architectures for primary and auxiliary modality transformations.
However, existing unified multimodal tracking frameworks often perform coarse multi-level interactions on all modality features within the encoder, inevitably introducing \textit{irrelevant background noise into the search area}. 
In this work, we propose a novel BSI module to leverage the correlation strength among the template, temporal information, and search area to \textit{emphasize target features while suppressing background interference}.



\noindent \textbf{Temperoal Modeling}.
Temporal information is crucial for tracking models to capture long-term changes and motion trends of the target. 
In RGB-based object tracking, researchers~\cite{stark,seqtrack,swintrack,cttrack,mixformer} have carefully designed various update strategies, typically guiding current state tracking through the fusion of accumulated templates. 
STMTrack~\cite{stmtrack} proposed a spatial-temporal memory network to exploit historical information. 
In contrast, multimodal tracking scenarios are more complex, requiring the consideration of not only RGB information but also the integration of additional modalities. 
In RGB-T Tracker, DMSTM~\cite{DMSTM} uses a dual-modality space-time memory network to aggregate historical information as well as the apparent information of the current frame. 
TATrack~\cite{tatrack} have successfully improved performance by combining dynamic template frames of the two modalities. 

However, exploration in the temporal dimension of multimodal tracking currently faces two main challenges:
$1$) Updating reference materials, such as template images and historical frame information, often depends on preset conditions, leading to \textit{sparse temporal relationships}. 
This limitation disrupts the continuous flow of information, making it difficult for the model to accurately capture ongoing target movements and changes over time.
$2$) Existing temporal exploration designs primarily focus on \textit{single multimodal task}, limiting their effectiveness in multi-task environments.
In contrast, our STTrack framework leverages explicit frame-to-frame temporal information exchange, capturing the target's temporal evolution using video context. 
Our method improves \textit{temporal continuity and contextual coherence}, as verified in the experiment section, and demonstrates its potential for unified application across \textit{various visual multimodal tasks}.




\section{Methodology}
In this paper, we introduce a novel spatial-temporal tracker (STTrack) based on temporal information, enabling continuous frame-to-frame information transfer through spatial-temporal data. 
Fig.~\ref{fig:structure} illustrates the overall architecture of STTrack. In this section, we first briefly review the state space model. Subsequently, we provide a detailed introduction to the overall architecture of our STTrack.

\subsection{Preliminaries}
\label{sec:pre}
The state space models draw inspiration from continuous linear time-invariant (LTI) systems.
The aim of SSM is to transform a one-dimensional function or sequence, represented as $x(t)$, into $y(t)$ via the hidden space $h(t) \in R^{N}$ with linear complexity. The system can be represented mathematically by the following formula:
\begin{equation}
\begin{gathered}
h^{\prime}(t) = Ah(t) + Bx(t),\\
  y(t) = Ch(t) + Dx(t),
\end{gathered}
  \label{eq:ssm}
\end{equation}
where the system's count parameters include the evolution parameter
$A \in \mathbb{R}^{N \times N}$, projection parameters $B \in \mathbb{R}^{N \times 1}$ and $C \in \mathbb{R}^{1 \times N} $, and skip connection $D\in\mathbb{R}$.
The $h^{\prime}(t)$ refers to the time derivative of $h(t)$, and $N$ is the state size.

When handling discrete sequences such as images and text, state space models need to convert continuous-time signals into discrete-time signals to accommodate the nature of discrete data. SSM adopt zero-order hold (ZOH) discretization to map the input sequence $\{x^1, x^2, ..., x^k\}$ to the output sequence $\{y^1, y^2, ..., y^k\}$. Specifically, suppose $\mathrm{\Delta}$ as the pre-defined timescale parameter to transformer continuous parameters $A$, $B$ to discrete space $\overline{A}$, $\overline{B}$. The discretization process is defined as follows:

\begin{equation}
\begin{gathered}
\overline{A}=\exp (\mathrm{\Delta} A), \\
\overline{B} = (\mathrm{\Delta} A)^{-1}(\exp (A) - I) \cdot \mathrm{\Delta} B.
  \end{gathered}
\end{equation}
After the discreization, Eq.~(\ref{eq:ssm}) can be rewritten as:
\begin{equation}
\begin{gathered}
h^{k} = \overline{A}{h^{k-1}} +\overline{B}{x^k}, \\
y^k = C{h^k} + D{x^k}.
  \end{gathered}
\end{equation}

SSM excels at modeling discrete sequences, but their inherent LTI property results in fixed parameters, making them insensitive to input variations. To overcome this limitation, a novel approach called the Selective State Space Model, also referred to as Mamba~\cite{mamba,videomamba,panmamba,simba,panmamba}, has been introduced.
Mamba makes model parameters dependent on the input data. It derives matrices $B$, $C$, and $\mathrm{\Delta}$ directly from the input $x$, allowing the model to adapt to different contexts and capture complex interactions within long sequences.


%


%

\subsection{Tracking Process} 
The multimodal tracking task generally involves integrating two distinct video modalities, which collaboratively contribute to the final decision-making process for tracking objects. For the input, each modality's data is first converted into the corresponding template tokens ($Z_{RGB}, Z_{X}$) and search tokens ($S_{RGB}, S_{X}$) through patch embedding and positional embedding encoding.
These tokens are then concatenated with the temporal information tokens that generated from the previous time state and fed into the tracker together. As shown in Fig.~\ref{fig:contrast}~(c), STTrack constructs a bridge between spatial and temporal information through its architecture, which consists of \textit{a visual encoder, a temporal state generator, a mamba fusion module, and a prediction head}.
We employ ViT~\cite{vit_transformer} as the visual encoder, with shared weights across the encoders, and insert background suppression interactive modules after each transformer layer. 
The visual encoder dynamically extracts precise multimodal features from the input multimodal images and prior temporal information. 
These features are fed into the temporal state generator to produce the current temporal information tokens, which are then passed to the next time point. The tracker then refines and fuses the visual features in the mamba fusion module, ultimately delivering them to the prediction head for the final tracking results.

\subsection{Temporal State Generator}
Previous methods typically focused on multimodal spatial features to achieve precise tracking results. 
However, these trackers are less effective in addressing challenges such as \textit{changes in moving targets and interference from similar objects}. 
To better capture target changes, it is crucial to construct stable inter-frame information features.
We introduce a temporal state generator that merges the unidirectional recurrent approach of cross mamba, employing autoregressive modeling to seamlessly transfer information from previous time nodes to the current one. 
This process integrates current multimodal spatial information to generate the temporal information tokens $T^{cur}_{RGB}$ and $T^{cur}_{X}$.
Specifically, the temporal state generator takes features from two modalities $ x_{RGB} = [Z_{RGB};S_{RGB};T^{pre}_{RGB}] $ and  $ x_{X} = [Z_{X};S_{X};T^{pre}_{X}] $
as input, generating the target state for the current time node and multimodal features after modality interaction. Where $T^{pre}$ is the temporal information learned from the previous $m$ frames.
Notably, at this stage, an empty token as $T^{cur}$ was inserted at the end to store the target information at the current time node.
We first apply $1$D convolution to $x_{RGB}$ and $x_{X}$, then linearly project them to produce the features $\overline{B}$, $\overline{C}$, and $D$ as described in the preliminaries.
By exchanging the $C$ matrix, the temporal state generator can incorporate complementary information from another modality when generating the current temporal token. 
Specifically, this process can be represented as:

\begin{equation}
\begin{gathered}
  h_{RGB}^{k} = \overline{A}_{RGB}{h_{RGB}^{k-1}} +\overline{B}_{RGB}{x_{RGB}^k},
\\
  y_{RGB}^k = \boldsymbol{{C}_{X}}{h_{RGB}^k} + {D}_{RGB}{x_{RGB}^k}.
  \end{gathered}
\end{equation}

\begin{equation}
\begin{gathered}
\ h_{X}^{k} = \overline{A}_{X}{h_{X}^{k-1}} +\overline{B}_{X}{x_{X}^k}, \\
  \ y_{X}^k = \boldsymbol{{C}_{RGB}}{h_{X}^k} + {D}_{X}{x_{X}^k},
  \end{gathered}
\end{equation}
where $k \in [1,2,..,l]$, $l$ is the length of visual tokens, and $y^k_{RGB}, y^k_{X}$ are concatenated to generate the visual features $y^{}_{RGB}, y^{}_{X}$.  The original mamba block is designed for the 1D sequence, which limits their ability to understand visual tokens with spatial location information.
Therefore, we adopt the commonly used bidirectional scanning method~\cite{vimamba} in visual Mamba to process the visual tokens. Specifically, we reverse the order of the visual tokens and perform calculations, then add the results of the reversed calculations to those of the non-reversed calculations. 

Extracting the current state information using the temporal information token allows us to add it to the queue $T$ and propagate it to the next frame:
\begin{equation}
\begin{gathered}
    T = \begin{cases}
   [T_1,..,T_{t-1},T_t] &\text{if } t<m  \\
   [T_{t-m},..,T_{t-1},T_t] &\text{if } t>=m. 
\end{cases} \\
  \end{gathered}
\end{equation}
where $m$ is number of temporal information tokens, and $t$ is time node. Temporal information tokens act as a bridge, linking the past, present, and future by using previous data to guide future modeling. 
Replacing the $C$ matrix with cross-modal attention allows our temporal tokens to capture more comprehensive information.

\begin{figure}[t]
  \centering
    \includegraphics[width=1\linewidth,height=6.2cm]
    {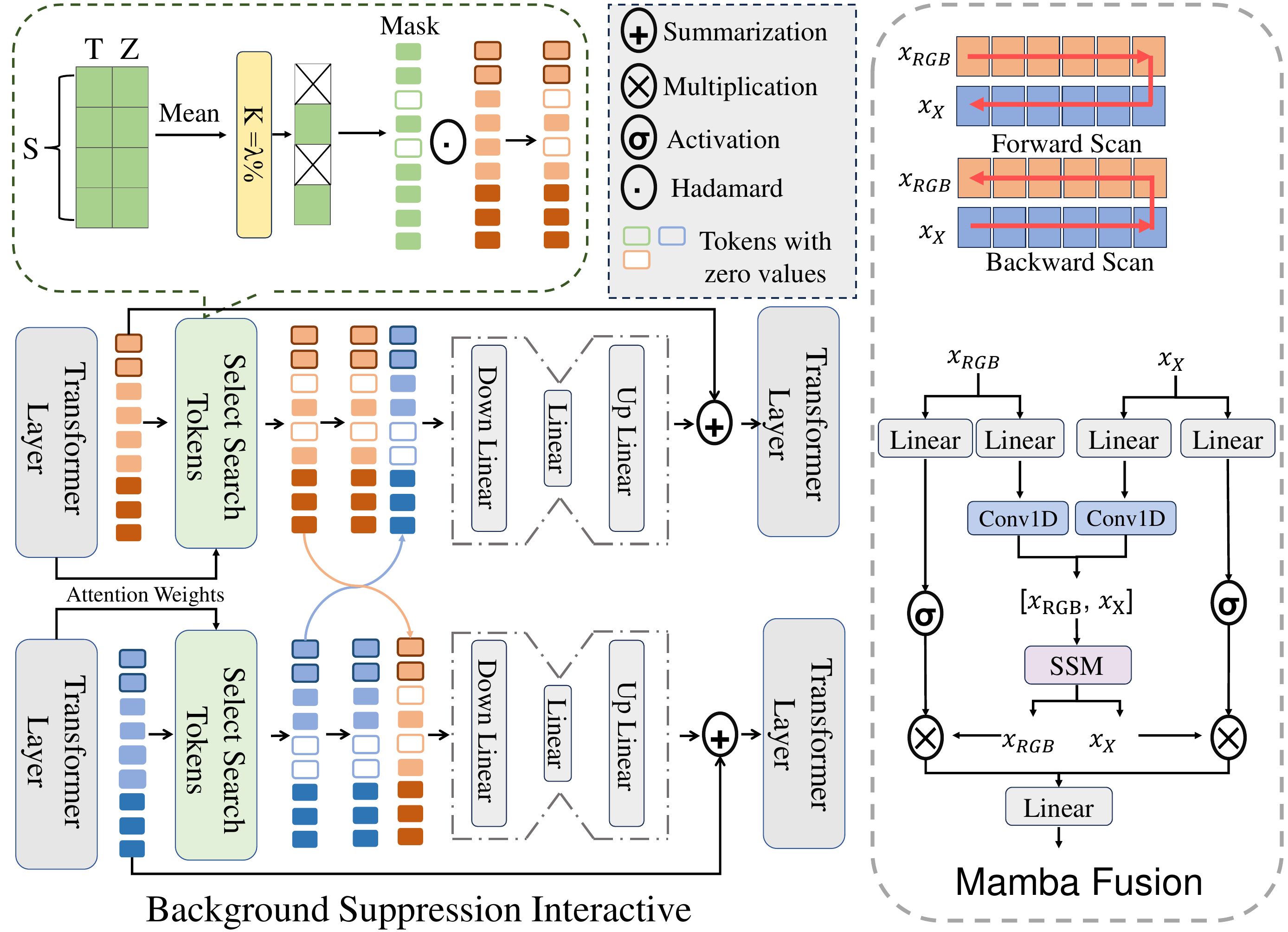}
   \caption{
\textbf{Left}: Architecture of the background suppression interactive module. 
\textbf{Right}: Details of the fusion mamba. 
In BSI module S is a search areas tokens, Z denotes the template tokens and T is the temporal information tokens. }
   \label{fig:bei_and_fusion}
\end{figure}

\subsection{Background Suppression Interactive}

%
%
Incorporating multiple interactions within the encoder has become the mainstream method in multimodal tracking. 
To this end, we incorperate our background suppression interactive (BSI) module to each layer of the encoder to enhance cross-modal interactions. 
Our visual encoder retains ViT architecture, where its self-attention~\cite{attention} mechanism is generally regarded as spatial aggregation of normalized tokens. 
Therefore, the similarity between tokens can be captured by the attention map, calculated as follows:
\begin{equation}
\begin{gathered}
W_{Z} = \text{softmax}( \frac{Q_Z \times K_S}{\sqrt{d}} ), \\
W_{T} = \text{softmax}( \frac{Q_T \times K_S}{\sqrt{d}} ),
  \end{gathered}
\end{equation}
where $W_{Z}$ represents the correlation between the search area and the template features, while $W_{T}$ represents the correlation between the search area and temporal information.

We use the attention computed in ViT as the criterion for background suppression, thereby avoiding additional computations. 
When calculating similarity with the search area, we use a $3 \times 3$ matrix centered on the template along with temporal information tokens to compute and average the results. 
Since tracking is essentially a matching task, a low similarity between search area tokens and the template region likely indicates a background area. 
Our temporal information tokens provide sufficient guidance to represent the target. 
By setting the filtering ratio $\lambda$, we first sort the search tokens by their similarity. \textit{Then, we select the bottom $\lambda$ proportion of tokens with the lowest similarity, mark them as invalid, and set their values to zero.}

Moreover, with each iteration of feature modeling by the visual encoder, the accuracy of the generated association matrix improves progressively.
Therefore, we divide the filtration ratio of the 12 layer BSI into three parts and gradually increase the $\lambda$.
As in Fig.~\ref{fig:bei_and_fusion}, after background suppression, we concatenate the features from the two modalities and generate cross-modal feature prompts through linear layers:
\begin{equation}
\begin{gathered}
x^i_{RGB} = F_{RGB}^i([f_{RGB};f_{X}]), \\
x^i_{X} = F_{X}^i([f_{X};f_{RGB}]),
  \end{gathered}
\end{equation}
where $f_{RGB}$, $f_{X}$ represent the features after background suppression, and $i$ denotes the transformer layer number. 



\subsection{Mamba Fusion}
Mamba excels in long-sequence modeling capabilities. Building on this, we concatenate the sequences of the two modalities and use a bidirectional scanning strategy to capture long-range dependencies in both modalities. Finally, we sum the two modality sequences to complete the modality fusion. The process can be represented as:
\begin{equation}
\begin{gathered}
    x = Fusion([Z_{RGB},S_{RGB}],[Z_{X},S_{X}]).
  \end{gathered}
\end{equation}
After obtaining $x_{\text{RGB}}\in R^{N\times C}$ and $x_{X}\in R^{N\times C}$, we concatenate them along the channel dimension and use a linear layer to adjust the dimension to $C$. 
Here $N$ represents the number of tokens of the feature sequence and $C$ represents the channel dimension. 
Detailed architecture is shown in Fig.~\ref{fig:bei_and_fusion}. 
In this way, we refine and fuse the features before they are fed into the prediction head.

\subsection{Head and Objective Loss}
Following most of the latest multimodal tracking methods~\cite{vipt, untrack}, we employ a stacked set of Fully Convolutional Networks (FCNs) ~\cite{ostrack} to construct the prediction head. Notably, during the tracking process, we maintain a temporal tokens $T$ with a length of $m$.
We use $ L_{\text{cls}} $~\cite{focal_loss} to denote the weighted focal loss for classification. For bounding box regression, we adopt the generalized IoU loss $ L_{\text{iou}} $~\cite{giou} and the $ L_1$ loss. The overall loss function of STTrack is:
\begin{equation}
L =  L_{\text{cls}} + \alpha L_{\text{iou}} + \beta L_1,
\end{equation}
where $ \alpha = 2$ and $  \beta = 5 $, which are hyperparameters to balance the contributions of loss terms.

\section{Experiment}
In this section, we begin by detailing the experimental training procedures and the inference process of the proposed STTrack. Following this, we compare STTrack against other leading methods using various benchmark datasets. 

\subsection{Implementation Details}
\textbf{Training.}
We train on multiple multimodal tasks, including LasHeR for RGB-T tracking, VisEvent for RGB-E tracking, and DepthTrack for RGB-D tracking. For input data, we use two 128 $\times$ 128 template images and one 256 $\times$ 256 search image. The training was conducted on four NVIDIA Tesla A6000 GPUs over 15 epochs, with each epoch consisting of 60,000 sample pairs and a batch size of 32. AdamW~\cite{adamw} was employed as the optimizer, with an initial learning rate of $1\mathrm{e}{-5}$ for the ViT backbone and $1\mathrm{e}{-4}$ for other parameters. After 10 epochs, the learning rate was reduced by a factor of 10.

\noindent \textbf{Inference.} During inference, we maintain the same training setting, using two template frames (includes a fixed initial template frame and a dynamically updated template frame.). 
Temporal information is incrementally incorporated into the tracking process, frame by frame. 
The tracking speed, tested on a NVIDIA 4090 GPU, is approximately $35.5$ frames per second (FPS).

%
   

\begin{table}[t]\normalsize

  \centering

    \small
    \fontsize{7.5}{8}\selectfont
    \begin{tabular}{l|c|cc|cc}
    \toprule
    \multirow{2}*{Method} &\multirow{2}*{Source} & \multicolumn{2}{c|}{LasHeR}  & \multicolumn{2}{c}{RGBT234} \\
        \cline{3-6} 
 & &   SR~$\uparrow$ & PR~$\uparrow$ & MSR~$\uparrow$ &MPR~$\uparrow$ \\
  \midrule[0.5pt]
    STTrack &Ours  & \textbf{60.3}  & \textbf{76.0} &   \textbf{66.7}	& \textbf{89.8} \\
    GMMT &AAAI'24 & \underline{56.6}  & \underline{70.7}  & \underline{64.7}  &\underline{ 87.9} \\
    BAT  &AAAI'24 & 56.3  & 70.2  & 64.1  & 86.8 \\
    TBSI &CVPR'24 & 56.3  & 70.5  & 64.3  & 86.4 \\
    TATrack &AAAI'24 & 56.1  & 70.2  & 64.4  & 87.2 \\
    OneTracker &CVPR'24 & 53.8  & 67.2  & 64.2  & 85.7 \\
    Un-Track  & CVPR'24 &53.6  & 66.7  & 61.8  & 83.7 \\
    SDSTrack &CVPR'24 & 53.1  & 66.5 & 62.5  & 84.8 \\
    ViPT  &CVPR'23 & 52.5  & 65.1  & 61.7  & 83.5 \\
    OSTrack &ECCV'22 & 41.2  & 52.5  & 54.9  & 72.9 \\
     ProTrack &MM'22 & 42.0 &53.8     & 59.9  & 79.5 \\
    APFNet  &AAAI'23& 36.2  & 50.0    & 57.9  & 82.7 \\
    \bottomrule
    \end{tabular}
    \caption{Comparisons on \textbf{RGB-T tracking}.}
\label{tab-sota-rgbt}
  
\end{table}

\begin{figure*}
  \centering
    \includegraphics[width=1\linewidth,height= 6.5 cm]
    {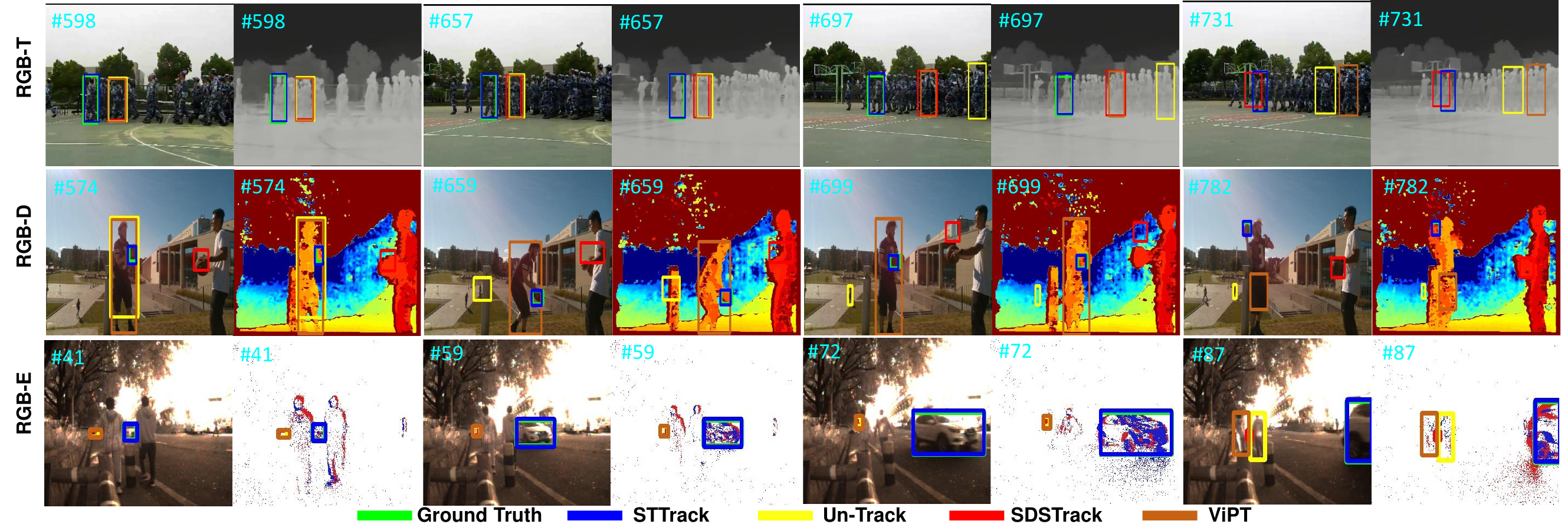}
   \caption{
Qualitative comparison between our method and other unified multimodal trackers on three multimodal task. The three sequences correspond to scenarios involving similar object interference, fast motion, and target deformation. Our tracker effectively addresses these challenges through dual optimization in both the temporal and spatial dimensions.}
   \label{fig:tracker_contrast}
\end{figure*}

\begin{table}[t]\normalsize

  \centering

      \small
    \fontsize{7}{8}\selectfont
    \begin{tabular}{l|ccc|ccc}
    \toprule
    \multirow{2}*{Method} & \multicolumn{3}{c|}{VOT-RGBD22}  & \multicolumn{3}{c}{DepthTrack} \\
        \cline{2-4} \cline{5-7}
 & EAO~$\uparrow$ & Acc.~$\uparrow$ & Rob.~$\uparrow$  &F-score~$\uparrow$ &Re~$\uparrow$ & Pr~$\uparrow$ \\
    \midrule[0.5pt]
    STTrack &  \textbf{77.6}     &   \textbf{82.5}    & \textbf{93.7}     & \textbf{63.3}  & \textbf{63.4}  & \textbf{63.2} \\
    SDSTrack & \underline{72.8}  & 81.2  & \underline{88.3}   & \underline{61.4}  & 60.9  & \underline{61.9} \\
    OneTracker & 72.7  & 81.9  & 87.2   & 60.9  & 60.4  & 60.7 \\
    Un-Track& 71.8  & \underline{82.0}    & 86.4   & 61.2    & \underline{61.0} & 61.3 \\
    ViPT  & 72.1  & 81.5  & 87.1    & 59.4  & 59.6  & 59.2 \\
    SBT-RGBD & 70.8  & 80.9  & 86.4   & -      & -      & - \\
    OSTrack & 67.6  & 80.3  & 83.3   & 52.9  & 52.2  & 53.6 \\
    DET   & 65.7  & 76.0    & 84.5   & 53.2  & 50.6  & 56.0 \\
    ProTrack & 65.1  & 80.1  & 80.2   & 57.8  & 57.3  & 58.3 \\
    SPT   & 65.1  & 79.8  & 85.1    & 57.8  & 53.8  & 52.7 \\
    STARK-RGBD & 64.7  & 80.3  & 79.8   &-       & -      & - \\
    KeepTrack & 60.6  & 75.3  & 73.9   &  -     &   -    & - \\
    ATCAIS & 55.9  & 76.1  & 73.9  & 47.6  & 45.5  & 50.0 \\
    \bottomrule
    \end{tabular}
    
 \caption{Comparisons on \textbf{RGB-Depth tracking}.}
 \label{tab-sota-rgbd}
\end{table}

   \begin{table*}[t]
    \centering
    \fontsize{6.5}{8}\selectfont

    \begin{tabular}{c|ccccccccccc|c}
     \toprule
   &STARK\_E &PrDiMP\_E &LTMU\_E &ProTrack &TransT\_E 
   &SiamRCNN\_E &OSTrack &Un-Track &ViPT & SDSTrack &OneTrack &STTrack\\ 
    \toprule
   AUC $\uparrow$ 
   &44.6 &45.3 &45.9 &47.1 &47.4 &49.9 &53.4 &58.9 &59.2 &59.7 &\underline{60.8} &\textbf{61.9} \\
  Pr  $\uparrow$
  &61.2&64.4&65.9 &63.2 &65.0 &65.9 &69.5 &75.5 &75.8 &\underline{76.7} &\underline{76.7} &\textbf{78.6}\\
       
    \bottomrule
    \end{tabular}
    \caption{Comparisons on \textbf{RGB-Event tracking}.}

    \label{tab-sota-rgbe}
    \end{table*}



  

\begin{figure}[t]
  \centering
    \includegraphics[width=1\linewidth,height=4.8cm]
    {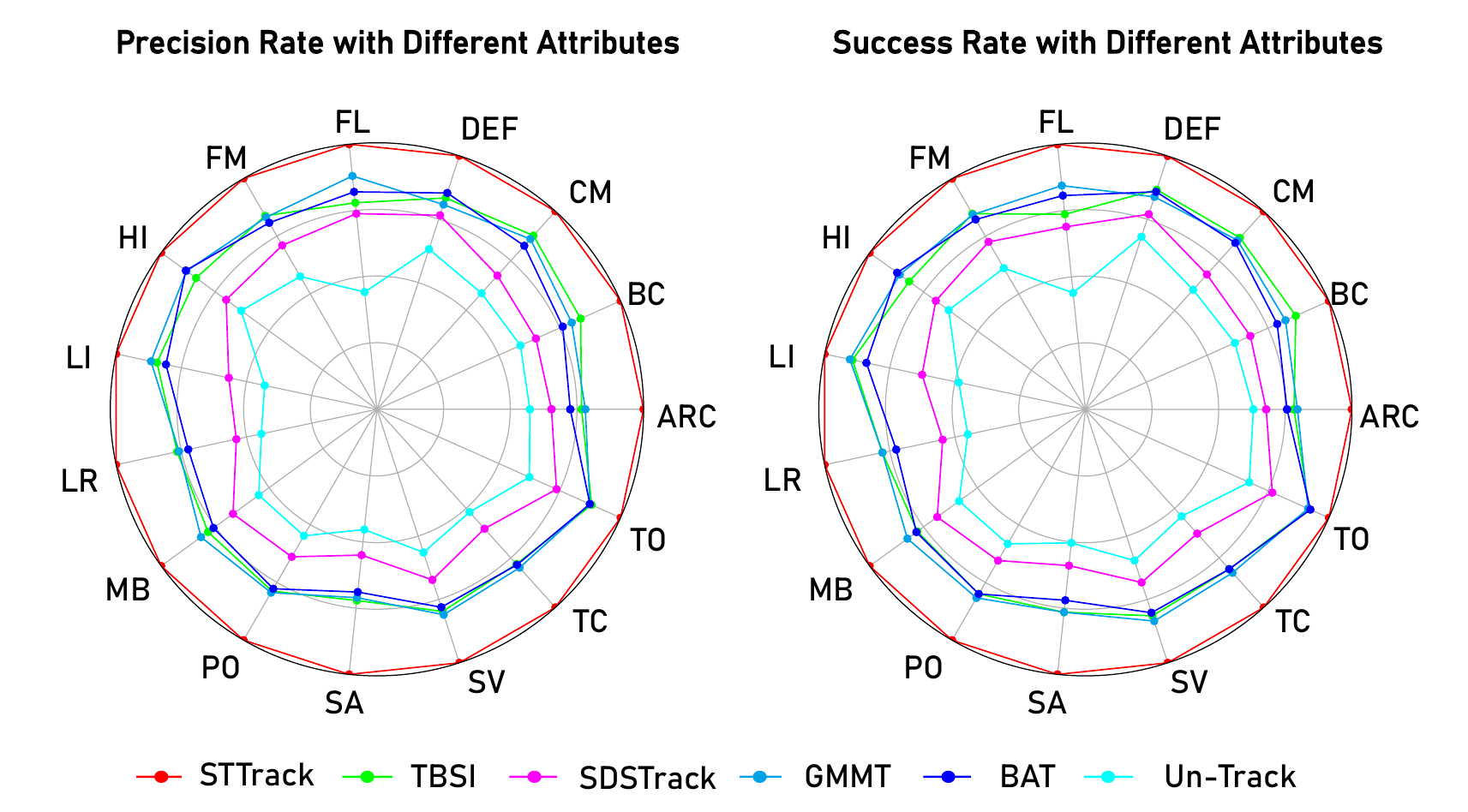}
   \caption{
Comparison of STTrack and SOTA trackers (including unified trackers and RGB-T trackers) under different attributes in the LasHeR dataset. }
   \label{fig:attribute}
\end{figure}

 \subsection{Comparison with State-of-the-Arts}
 
 \textbf{LasHeR}. The LasHeR represents a substantial RGB-T tracking dataset comprising 1224 aligned sequences, encompassing more than 2$\times$730K frames captured across diverse imaging platforms. 
As shown in Tab.~\ref{tab-sota-rgbt}, 
our STTrack achieved an SR of 60.3\% and a PR of 76.0\%, surpassing GMMT by 4.3\% in SR and 5.3\% in PR, demonstrating the effectiveness of continuous spatial-temporal modeling.
 
 
\noindent \textbf{RGBT234.} The RGBT234 benchmark introduces enriched annotations and an expanded set of environmental challenges. It contains 234 aligned sequences of RGBT videos. As illustrated in Tab.~\ref{tab-sota-rgbt}, STTrack achieves the best MSR score of 66.7\%, outperforming the recent trackers.

\noindent \textbf{DepthTrack}. DepthTrack is a long-time tracking dataset in which the average sequence length is 1,473 frames. 
The dataset covers 200 sequences, 40 scenes and 90 target objects. As in Tab.~\ref{tab-sota-rgbd}, our STTrack obtains SOTA results with 63.3\% in F-score, 63.4\% in recall, and 63.2\% in precision.

\noindent \textbf{VOT-RGBD2022}. VOT-RGBD2022 comprises 127 brief RGB-D sequences and evaluates tracker performance using Accuracy, Robustness, and Expected Average Overlap (EAO) metrics. As illustrated in Tab.~\ref{tab-sota-rgbd}, our proposed tracker, STTrack, demonstrates a notable improvement in EAO, achieving a 4.8\% increase compared to the previous SOTA tracker SDSTrack.

\noindent \textbf{VisEvent}. VisEventis the largest RGB-E dataset currently available, encompassing 500 training video sequences and 320 testing video sequences. As reported in Tab.~\ref{tab-sota-rgbe}, our STTrack obtains SOTA AUC and precession of 61.9\% and 78.6\%, respectively. 

\begin{table}[t]
\centering

\small

\fontsize{7}{8}\selectfont
\begin{tabular}{l|c|ccc|c}
\toprule
\# & Method & LasHeR & DepThTrack  & Visevent & $\Delta$ \\
\midrule 
1 & Baseline & 56.0 & 58.8 & 60.0 & -- \\
2 & + Template Updata  & 57.1 & 59.5 & 59.8 & \textbf{+0.5} \\  
3 & + Temporal Information  & 58.9 & 62.0 & 61.1 & \textbf{+1.8} \\
4 & + Mamba Fusion  & 59.2 & 62.1 &61.3  & \textbf{+0.2}\\
5 & + BSI Module  & 60.3 & 63.2 &61.9 & \textbf{+0.9} \\
\bottomrule
\end{tabular}
\caption{Quantitative comparison among different variants of STTrack on the LasHeR dataset, DepThTrack dataset and Visevent dataset. `$\Delta$' denotes the performance change (averaged over benchmarks) compared with previous variants.
}
\label{tab:ablation}

\end{table}

\begin{table}[t]\normalsize
\centering

\fontsize{6}{6}\selectfont

     \small
    \begin{tabular}{c|cccc}
    \toprule
    Dataset &1 &2&\cellcolor{gray!15}4&8\\
    \midrule
    LasHeR  &59.6 &59.9 &\cellcolor{gray!15}60.3 &60.0\\
    DepthTrack
    &61.0  &61.2  &\cellcolor{gray!15}63.2 &62.9\\
    VisEvent &61.4 &61.6 &\cellcolor{gray!15}61.9 &61.7\\
    $\Delta$ &-- &+0.2 & \cellcolor{gray!15}+0.9 &-0.3\\  
    \bottomrule
    \end{tabular}

  \caption{Ablation study on the number of temporal information tokens. We use \textcolor{gray}{gray} color to denote our final trackers setting. '$\Delta$' denotes the performance change (averaged over benchmarks) compared with previous number setting.
  }
    \label{tab:ablation_token}
\end{table}

\begin{table}[t]
\centering

\small

\fontsize{8}{9}\selectfont
\begin{tabular}{l|c|ccc}
\toprule
\# &  Filtering Ratio ($\lambda\%$) & LasHeR & DepThTrack  & Visevent  \\
\midrule 
1 & [0\%,0\%,0\%] & 59.4 & 61.6 & 61.1  \\
2 & [15\%,15\%,15\%]  & 59.8 & 62.9 & 51.7  \\  
3 & [30\%,30\%,30\%]  & 59.6 & 62.1 & 61.5  \\
4 &\cellcolor{gray!15} [0\%,15\%,30\%]  & \cellcolor{gray!15}60.3& \cellcolor{gray!15}63.2 &\cellcolor{gray!15}61.9  \\
\bottomrule
\end{tabular}

\caption{Ablation study on the ration in BSI module. We use \textcolor{gray}{gray} color to denote our final trackers setting.
}
\label{tab:ablation_fil}

\end{table}

\subsection{Ablation Studies}

\textbf{Component Analysis}.
In Tab.~\ref{tab:ablation}, we conducted an ablation study using the AUC in LasHeR, the Precession in DepThTrack,and the AUC in Visevent. The baseline used ViT as the visual encoder and fused the two modalities through a convolutional layer before the prediction head to establish. Through experimental results, we found that template updates can timely refresh the target's appearance information, compensating for the initial template's shortcomings and significantly enhancing the model's tracking performance. The model's performance was not optimal because of its sparse template update method and lack of frame-to-frame information transfer. Therefore, the introduction of temporal information led to a significant improvement, resulting in a 1.8\% gain, which demonstrates the effectiveness of temporal information in addressing these issues. Furthermore, the results show that due to the differences in target representation across different modalities, optimizing the modality fusion process with mamba fusion module can further improve model performance. Additionally, by reducing interference from non-essential regions, our proposed background suppression scheme effectively enhances the performance of cross-modal interactions.

\noindent \textbf{Number of Temporal Information Tokens}. 
We investigate the impact of temporal information on the performance of STTrack, as shown in Tab.~\ref{tab:ablation_token}. As the number increases from 1 to 4, the model's performance improves, indicating the temporal information tokens positively contribute to optimization. 
However, when the number is increased further, performance declines, possibly due to earlier temporal information tokens failing to accurately describe the current target state, leading to the introduction of noise. Therefore, we selected an optimal number of temporal information tokens.

\noindent \textbf{Filtering Ratio in BSI}. 
To validate the impact of background suppression on performance, we conducted experiments with different background filtering ratios. 
Here, the 12-layer BSI module was processed in three stages. 
As Tab.~\ref{tab:ablation_fil} indicates that while a fixed background filtering ratio can enhance performance, a three-stage approach with progressively increased filtering ratios yields more significant improvements. This is because, within a single-stream structure, the features of the search area, guided by the template and temporal information tokens, need to progressively highlight the foreground target layer by layer.


\subsection{Exploration Study and Analysis}
\textbf{Attribute-based Performance.}
We analyze the performance of our method in various scenarios by evaluating it on different attributes of LasHeR dataset.
As shown in Fig.~\ref{fig:attribute}, STTrack surpasses previous state-of-the-art trackers on these attributes. This improvement is due to our approach's enhanced temporal information and complementary spatial features, allowing STTrack to maintain stable tracking even when the target undergoes changes. STTrack excels in scenarios requiring temporal information, such as Partial Occlusion (PO) and Deformation (DEF), as well as in conditions with significant modality imaging differences, such as low-light and high-light situations.



\begin{figure}[t]
  \centering
    \includegraphics[width=1\linewidth,height=3.2cm]
    {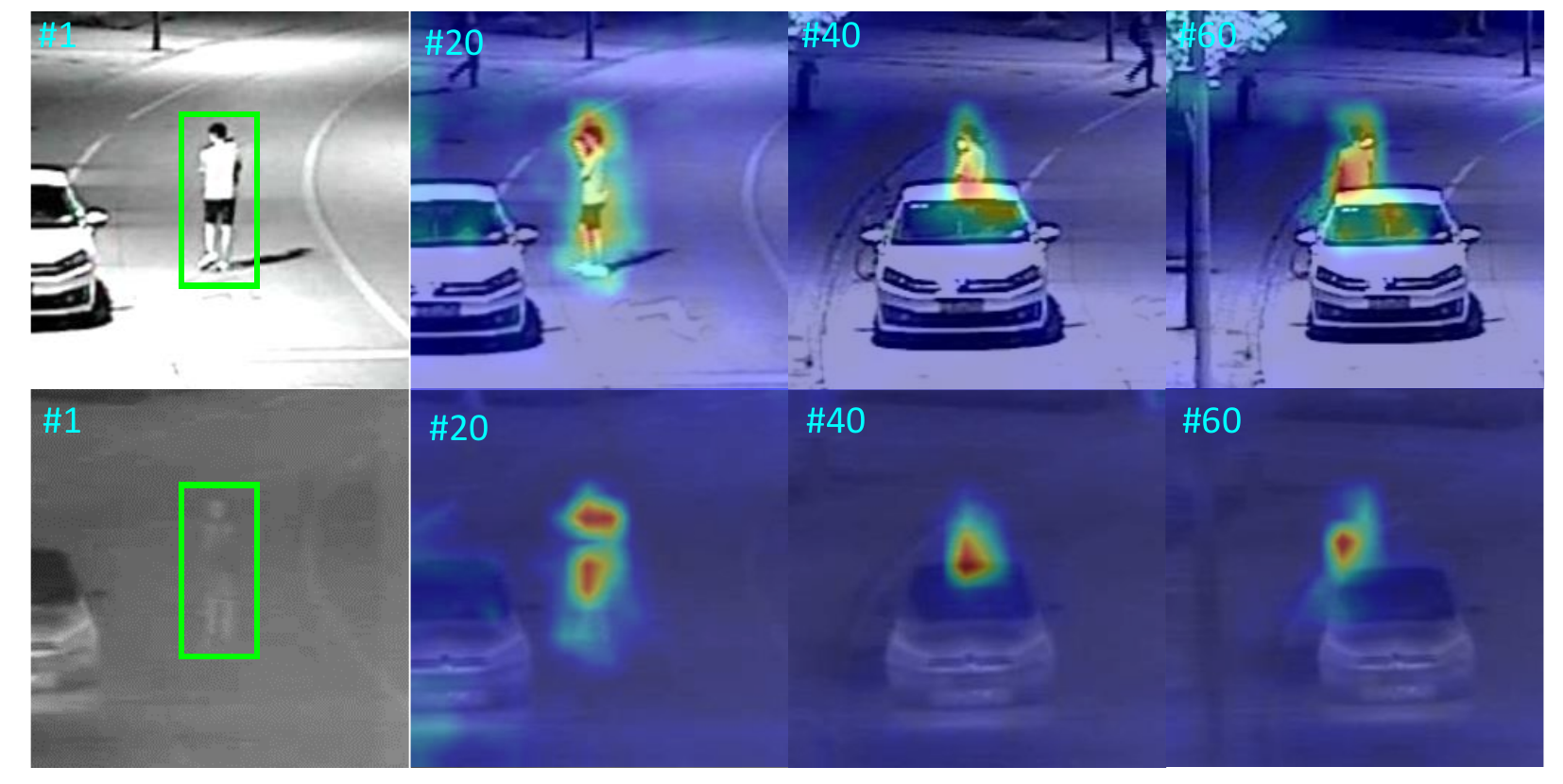}
   \caption{
 The attention map of temporal information tokens with search area. These visual results are in LasHeR.}
   \label{fig:attention}
\end{figure}

\noindent \textbf{Visualization Results}. 
As shown in Fig.~\ref{fig:tracker_contrast}, we qualitatively compare STTrack with three other multimodal unified trackers. In the RGB-T sequences, where similar objects cause significant interference and the target has clear movement directions, STTrack leverages temporal information for stable tracking. In the RGB-D sequences, despite severe occlusion, our method captures the target by utilizing the complementary strengths of the RGB and Depth modalities along with continuous temporal information. In the RGB-E sequences, where the car moves at high speed and undergoes significant deformation due to changes in camera distance, STTrack effectively tracks the target by gradually adapting to these changes over time. Besides, we visualize the attention map of the temporal information tokens with search area, as shown in Fig.~\ref{fig:attention}. It demonstrates that the continuous propagation of temporal markers and the focus on object temporal information can effectively capture and respond to the dynamic state of the target.

\section{Conclusion}
In this work, we propose a tracking framework named STTrack based on multimodal spatio-temporal patterns. By leveraging temporal context, STTrack effectively captures and represents dynamic targets. The tracker incorporates a temporal state generator to generate multimodal temporal information that supports the tracking process. Additionally, it is equipped with the BSI module and Mamba Fusion module, which optimize modality branch representation and fuse multimodal features at the spatial level. Compared to previous multimodal trackers, our approach achieves state-of-the-art performance across three multimodal tasks.

\section{Acknowledgements}
This work was funded by the National Science Fund of China, with Grant Nos. U24A20330, 62361166670, and 62406135, the Natural Science Foundation of Jiangsu Province under Grant No. BK20241198, and the AI \& AI for Science Project of Nanjing University, Grant No. 14380007.

\bibliography{aaai25}
\newpage
\twocolumn[{%
\renewcommand\twocolumn[1][]{#1}%
\begin{center}
    \centering
    \captionsetup{type=figure}
    \includegraphics[width=0.76\linewidth,height=6cm]{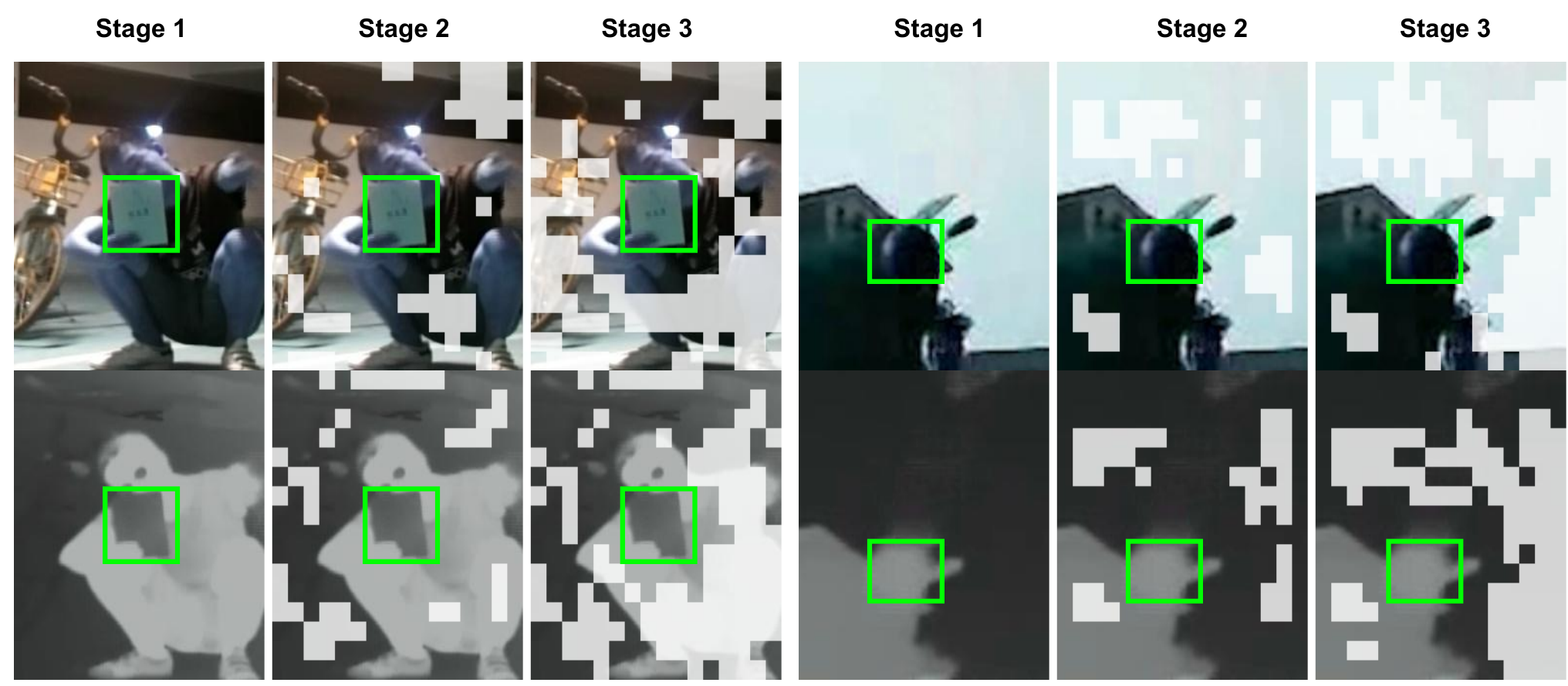}
    \captionof{figure}{
   Visualization of background suppression in the BSI module.
}
\label{fig:BSI}
    \includegraphics[width=0.8\linewidth]{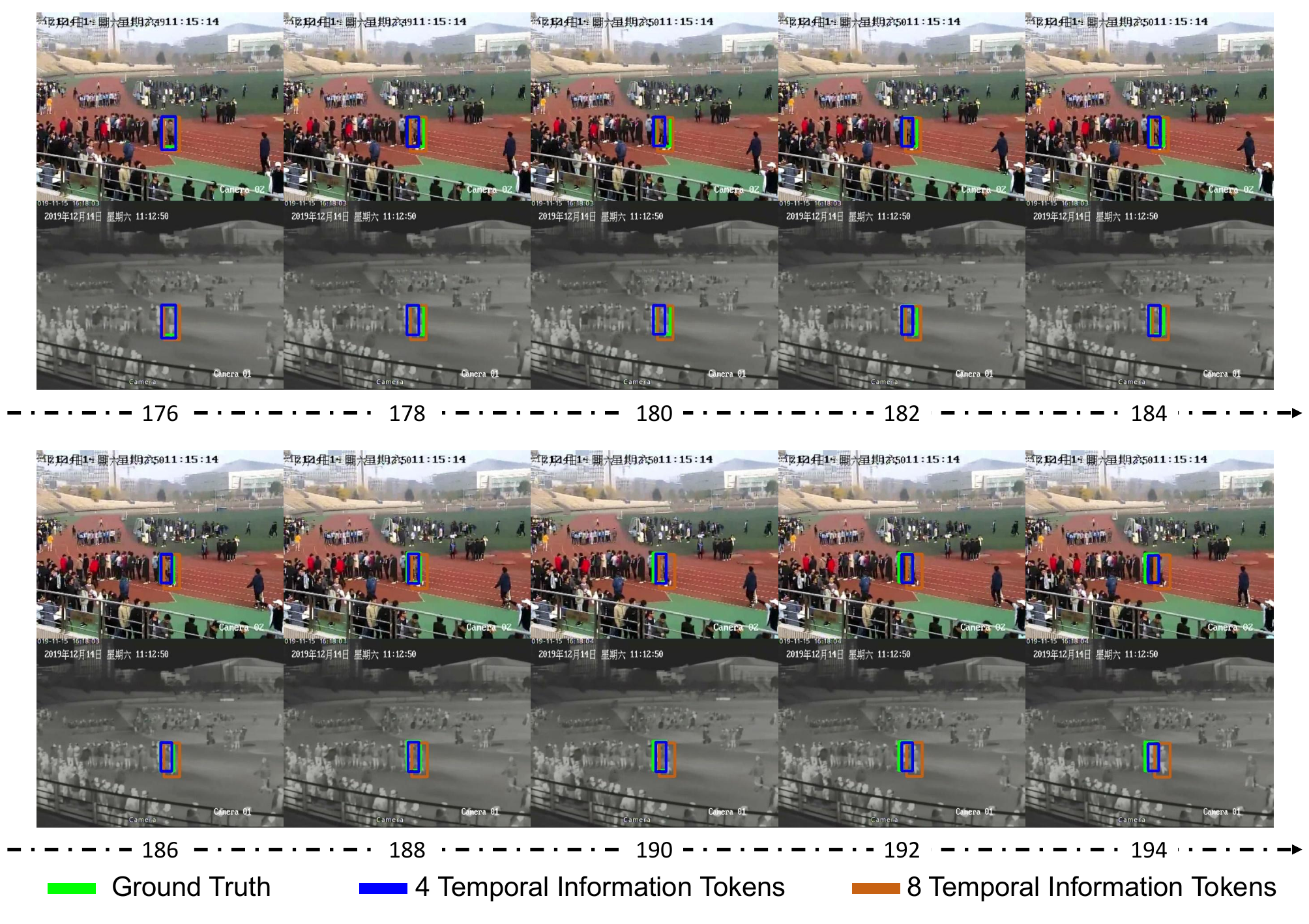}
    \captionof{figure}{
Visualization comparison of different quantities of temporal information tokens.
}
\label{fig:temporal}
\end{center}%
}]
\section{Supplementary Material}

\textbf{More discussion of background suppression}. To evaluate whether APTrack can suppress the background in BSI, we visualize this process. As shown in Fig.~\ref{fig:BSI}, BSI module effectively suppresses multimodal background interference by leveraging the correlation between each component and the search area. This mechanism not only reduces the impact of background noise but also accurately highlights target features, thereby enhancing the precision of interactions.

\noindent \textbf{More discussion of temporal information tokens
}. 
To explore why the performance dropped when the number of temporal information tokens increased from 4 to 8, we conducted a visual analysis of the two tracker versions. As shown in Fig.~\ref{fig:temporal}, when the number of tokens was 8, the tracker gradually lost track of the target in occlusion scenarios. This occurred because the more temporal information tokens introduced excessive redundant information, making it difficult for the model to effectively filter and focus on the current target state. In other words,  surplus temporal information tokens probably carried historical information that didn't align with the current target state, and this caused interference in the model's decision-making process.

\end{document}